\documentclass[11pt,a4paper]{article}
\usepackage[hyperref]{mrp20}
\usepackage{times}
\usepackage{latexsym}

\usepackage{microtype}

\aclfinalcopy 

\usepackage{amsmath} 
\usepackage{bm}
\usepackage{booktabs} 
\usepackage{colortbl} 
\usepackage{enumitem}
\usepackage{graphicx}  
\usepackage{multirow}
\usepackage{subcaption} 
\usepackage{tcolorbox} 
\usepackage{relsize}
\usepackage{mathtools} 
\usepackage{mymacros} 
\usepackage{txfonts} 
\usepackage{pifont}
\usepackage{marvosym}

\newcommand{\drgvar}[1]{\textbf{#1}}
\newcommand{\comm}[1]{\textcolor{gray!80}{#1}} 
\newcommand{\posn}[2]{#1\kern-0.15em.\kern-0.15em#2}
\newcommand{\graytxt}[1]{\textcolor{gray}{#1}}

\newcommand{\BBenc}{BB$^*$}
\newcommand{\Liu}[1]{L#1}

\newcommand{\bmB}{\ensuremath{\bm{B}}}
\newcommand{\bmC}{\ensuremath{\bm{C}}}
\newcommand{\bargs}{\ensuremath{A\text{\rotatebox{90}{\footnotesize{---\kern-6.5pt\textbullet}}}}}
\newcommand{\fargs}{\ensuremath{A_a^\mathsmaller{<}}}
\newcommand{\Cor}{\ensuremath{\bmC\kern-5.5pt\raisebox{1.8pt}{\smaller[2]\textbullet}\kern2.3pt}}
\newcommand{\Crln}{\ensuremath{\bmC^\circ}}
\newcommand{\Brln}{\ensuremath{\bmB^\circ}}
\newcommand{\Ctorn}{\ensuremath{\bmC^\text{\rotatebox{-90}{$\multimap$}}}}
\newcommand{\Btorn}{\ensuremath{\bmB^\text{\rotatebox{-90}{$\multimap$}}}}
\newcommand{\Ctor}{\ensuremath{\bmC^\text{\rotatebox{-90}{$\multimapdot$}}}}
\newcommand{\tacl}{chLSTM$^\uparrow$}
\newcommand{\poorLSTM}{chLSTM$^\downarrow$}
\newcommand{\Boxerup}{Boxer$^\uparrow$}
\newcommand{\Boxerdw}{Boxer$^\downarrow$}
\newcommand{\CorCut}{$\Cor$\kern.5pt\raisebox{-.8pt}{\rotatebox{90}{\footnotesize\ding{34}}}}

\newcommand{\gnode}[1]{%
\begin{tikzpicture}[baseline=(X.base)]
\tikzset{every node/.style={scale=0.6}}
\node (X) [draw, shape=circle, inner sep=1pt, outer sep=0] {#1\vphantom{$\bm{BC}$}};
\end{tikzpicture}}
    
\newcommand{\toreif}[1]{%
\begin{tikzpicture}[baseline=(X.base)]
\tikzset{every node/.style={scale=0.6}}
\node (X) [draw, shape=circle, inner sep=1pt] {\vphantom{$\bm{BC}$}};
\node (Y) at($(X)-(5mm,0)$) [inner sep=1pt] {};
\path [->,>=latex] (Y.east) edge[draw] node[pos=.3,sloped,above]{\small #1} (X.west); 
\end{tikzpicture}}

\newcommand{\edgetoref}[1]{%
\begin{tikzpicture}[baseline=(X.south)]
\tikzset{every node/.style={scale=0.6}}
\node (X) [draw, fill=black!70, shape=circle, inner sep=2.5pt]{};
\node (Y) at($(X)-(5mm,0)$) [inner sep=1pt] {};
\path [->,>=latex] (Y.east) edge[draw] node[pos=.3,sloped,above]{\small #1} (X.west); 
\end{tikzpicture}}

\newcommand{\rolmid}[2][0pt]{%
\begin{tikzpicture}[baseline=(Y.base)]
\tikzset{every node/.style={scale=0.6}}
\node (Y) [draw, shape=circle, inner sep=1pt] {#2\vphantom{$\bm{BC}$}};
\node (X) at($(Y)-(5.5mm,0)$) [inner sep=0pt] {1};
\node (Z) at($(Y)+(5.5mm,0)$) [inner sep=0pt] {2};
\path [->,>=latex] (X.east) edge[draw] node[pos=.3,sloped,inner sep=#1,shape=circle,fill=black,above]{} (Y.west); 
\path [->,>=latex] (Y.east) edge[draw] node[pos=.3,sloped,inner sep=#1,shape=circle,fill=black,above]{} (Z.west); 
\end{tikzpicture}}

\newcommand{\cnodeonref}[1]{%
\begin{tikzpicture}[baseline=(R.south)]
\tikzset{every node/.style={scale=0.6}}
\node (C) [draw, shape=circle, outer sep=0pt, inner sep=5pt, label={[label distance=0pt]180:#1}] {};
\node (R) at($(C)$)  [fill=black!70, shape=circle, inner sep=2pt]{};
\end{tikzpicture}}

\newcommand{\rolfork}[1]{%
\begin{tikzpicture}[baseline=(Y.base)]
\tikzset{every node/.style={scale=0.6}}
\node (Y) [draw, shape=circle, inner sep=1pt] {#1\vphantom{$\bm{BC}$}};
\node (X) at($(Y)+(7mm,1mm)$) [inner sep=0pt] {1};
\node (Z) at($(Y)+(7mm,-1mm)$) [inner sep=0pt] {2};
\path [->,>=latex] (Y.east) edge[draw] node[sloped,inner sep=1.4pt,shape=circle,fill=black,above,pos=.4]{} (X.west); 
\path [->,>=latex] (Y.east) edge[draw] node[sloped,inner sep=1.4pt,shape=circle,fill=black,below,pos=.4]{} (Z.west); 
\end{tikzpicture}}

\title{DRS at MRP~2020:\\
  Dressing up Discourse Representation Structures as Graphs}

\author{Lasha Abzianidze\thanks{\,\,Part of the work was done while the author was at the University of Groningen.}\\
  UiL OTS\\
  Utrecht University\\
  \texttt{l.abzianidze@uu.nl} \\
  \And
  Johan Bos\\
  CLCG\\
  University of Groningen\\
  \texttt{johan.bos@rug.nl} \\
  \And
  Stephan Oepen\\
  Department of Informatics\\
  University of Oslo\\
  \texttt{oe@ifi.uio.no } \\ }

\date{}

\begin{document}
\maketitle
\begin{abstract}
Discourse Representation Theory (DRT) is a formal account for representing the meaning of natural language discourse.
Meaning in DRT is modeled via a Discourse Representation Structure (DRS), a meaning representation with a model-theoretic interpretation, which is usually depicted as nested boxes.
In contrast, a directed labeled graph is a common data structure used to encode semantics of natural language texts.
The paper describes the procedure of dressing up DRSs as directed labeled graphs to include DRT as a new framework in the 2020 shared task on Cross-Framework and Cross-Lingual Meaning Representation Parsing.
Since one of the goals of the shared task is to encourage unified models for several semantic graph frameworks, the conversion procedure was biased towards making the DRT graph framework somewhat similar to other graph-based meaning representation frameworks.
\end{abstract}

\section{Introduction}\label{s:intro}
\begin{figure*}[t!]
\begin{subfigure}{.315\textwidth}
\mbox{\includegraphics[clip, trim=13.8mm 18mm 13.8mm 41mm, 
    width=\textwidth]{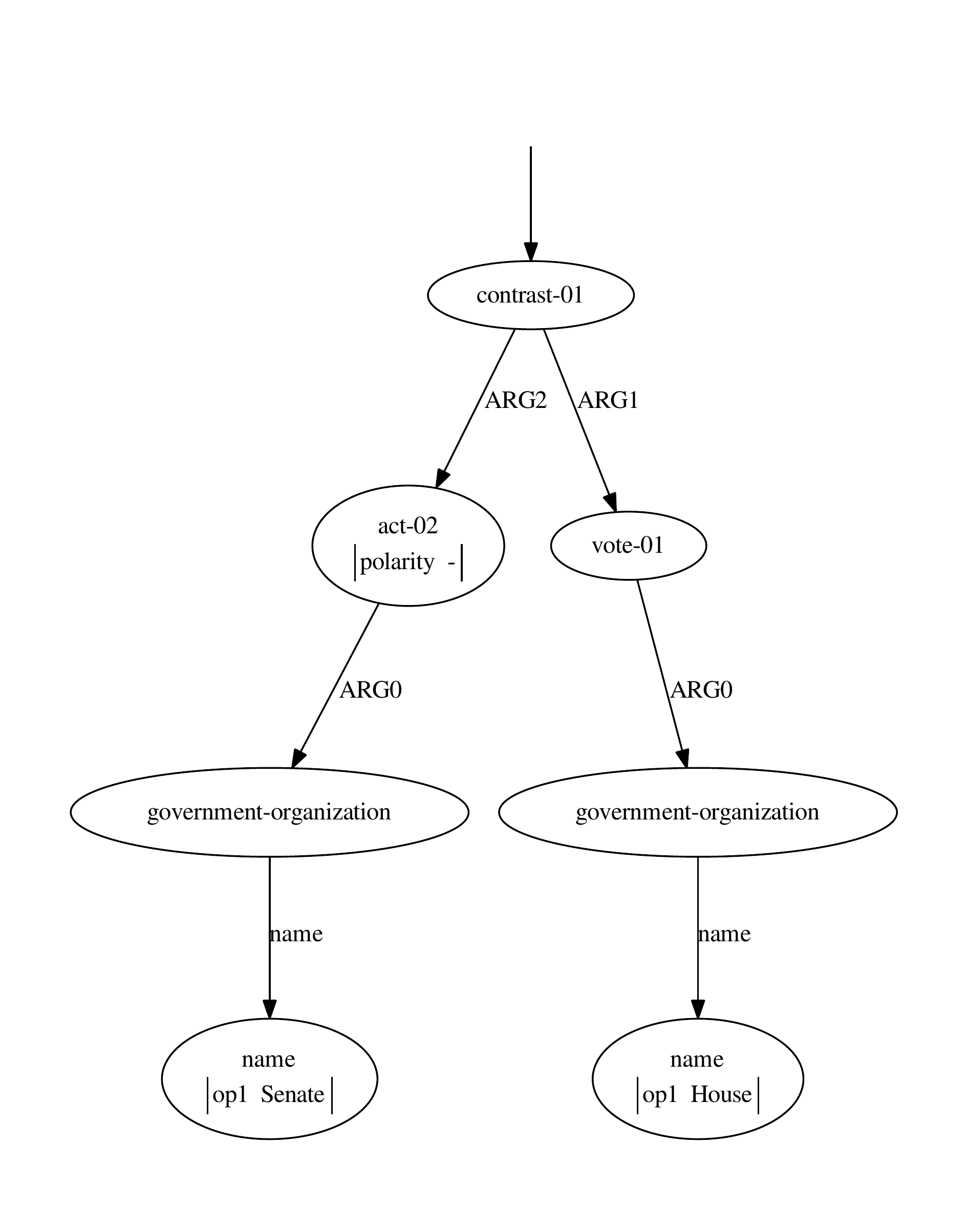}}
\caption{AMR: Abstract Meaning Representation \citep{banarescu-etal-2013-abstract} \phantom{012345678912345678}}
\label{sfig:amr}
\end{subfigure}
\hspace{5mm}
\begin{subfigure}{.255\textwidth}
\mbox{\includegraphics[clip, trim=14.7mm 14mm 14.7mm 29mm,
    width=\textwidth]{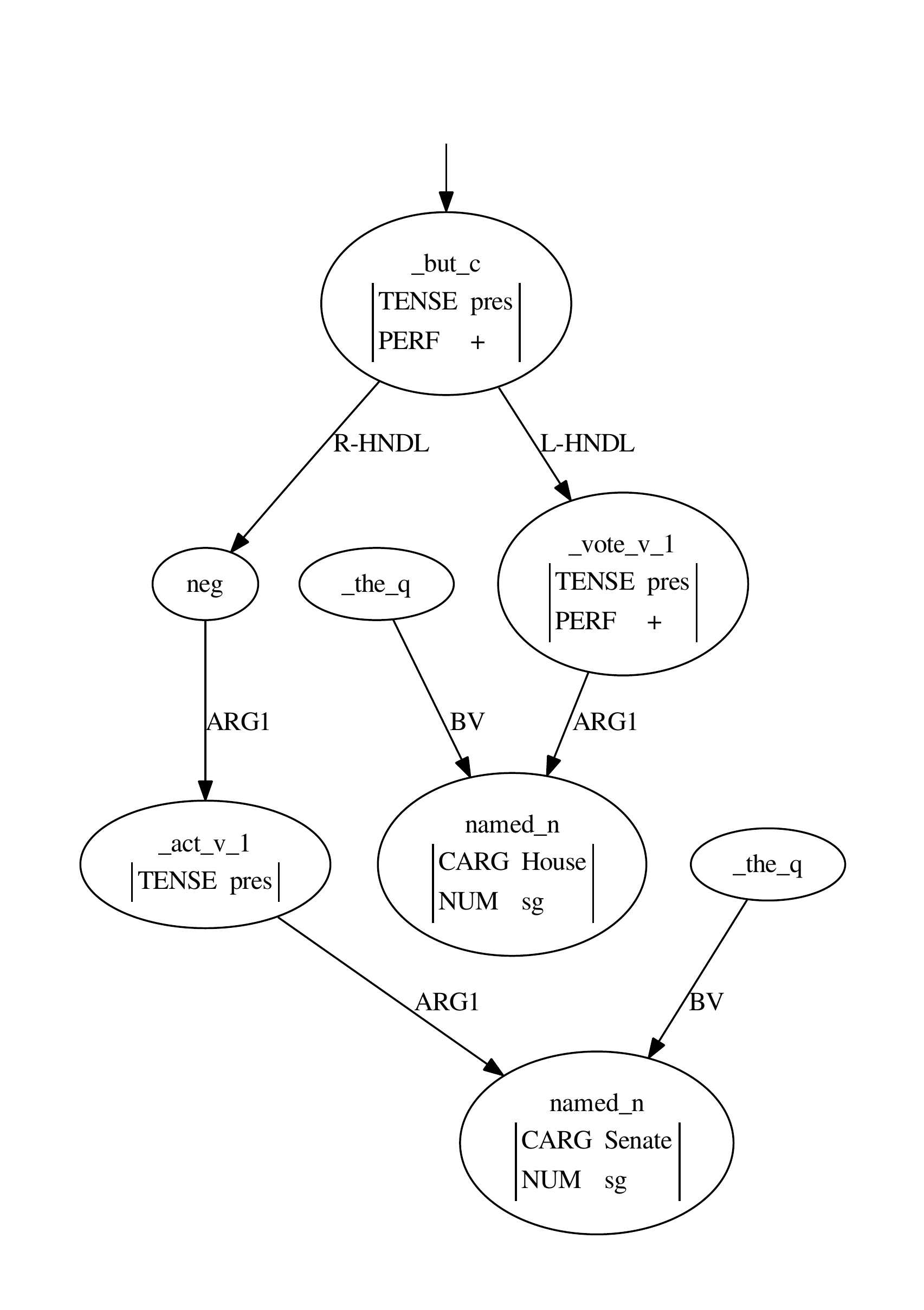}}
\caption{EDS: Elementary Dependency Structures \citep{oepen-lonning-2006-discriminant}}
\label{sfig:eds}
\end{subfigure}
\hspace{5mm}
\begin{subfigure}{.36\textwidth}
\mbox{\includegraphics[clip, trim=14mm 15mm 14.3mm 32mm,
    width=\textwidth]{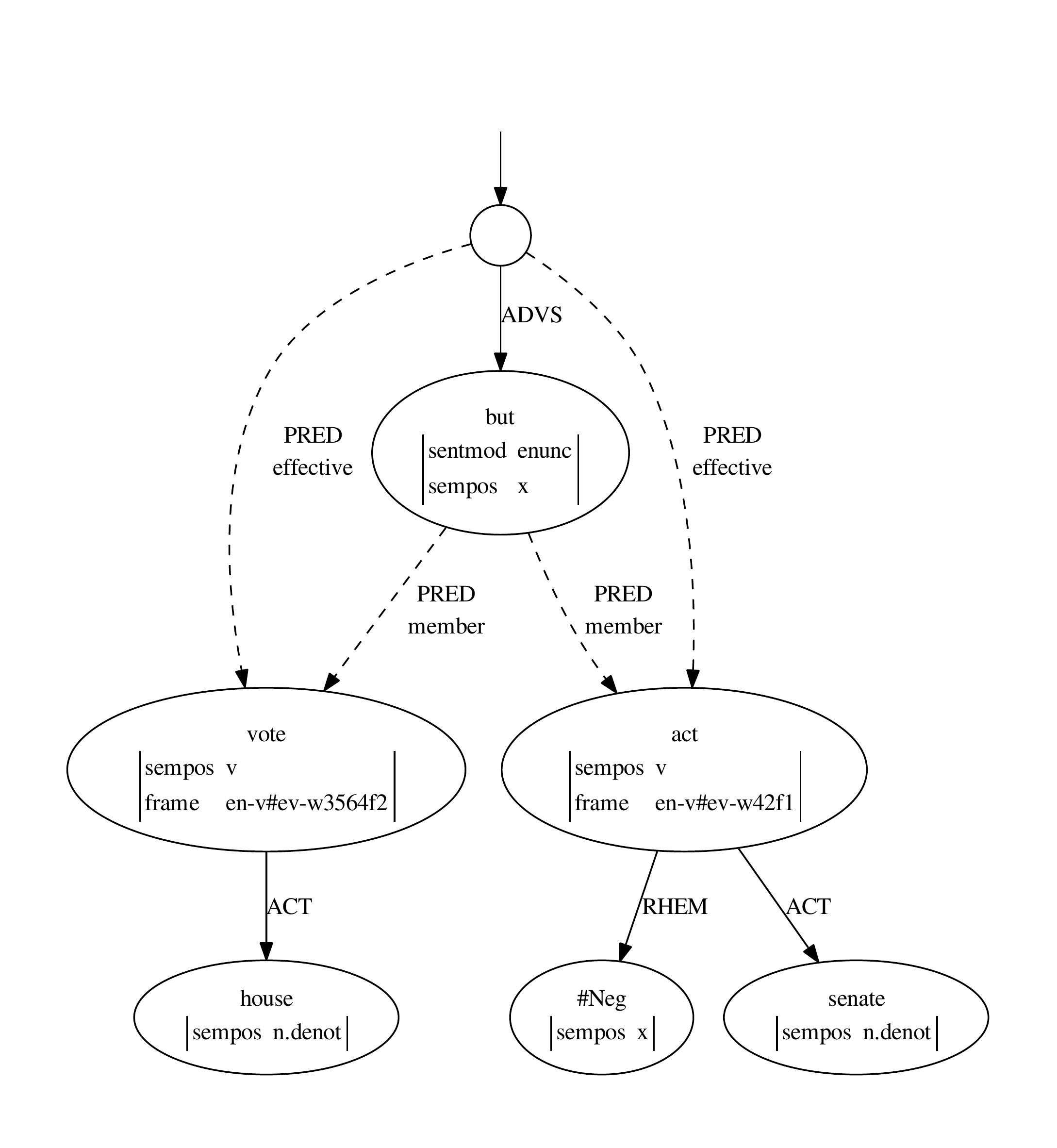}}
\caption{PTG: Prague Tectogrammatical Graphs \citep{sgallhp:1986,hajic-etal-2012-announcing,Zem:Haj:20} \phantom{012345678912345678}}
\label{sfig:ptg}
\end{subfigure}
\\\vspace{4mm}

\begin{subfigure}{.45\textwidth}
\mbox{\includegraphics[clip, trim=16.2mm 9mm 16.6mm 32mm,
    width=\textwidth]{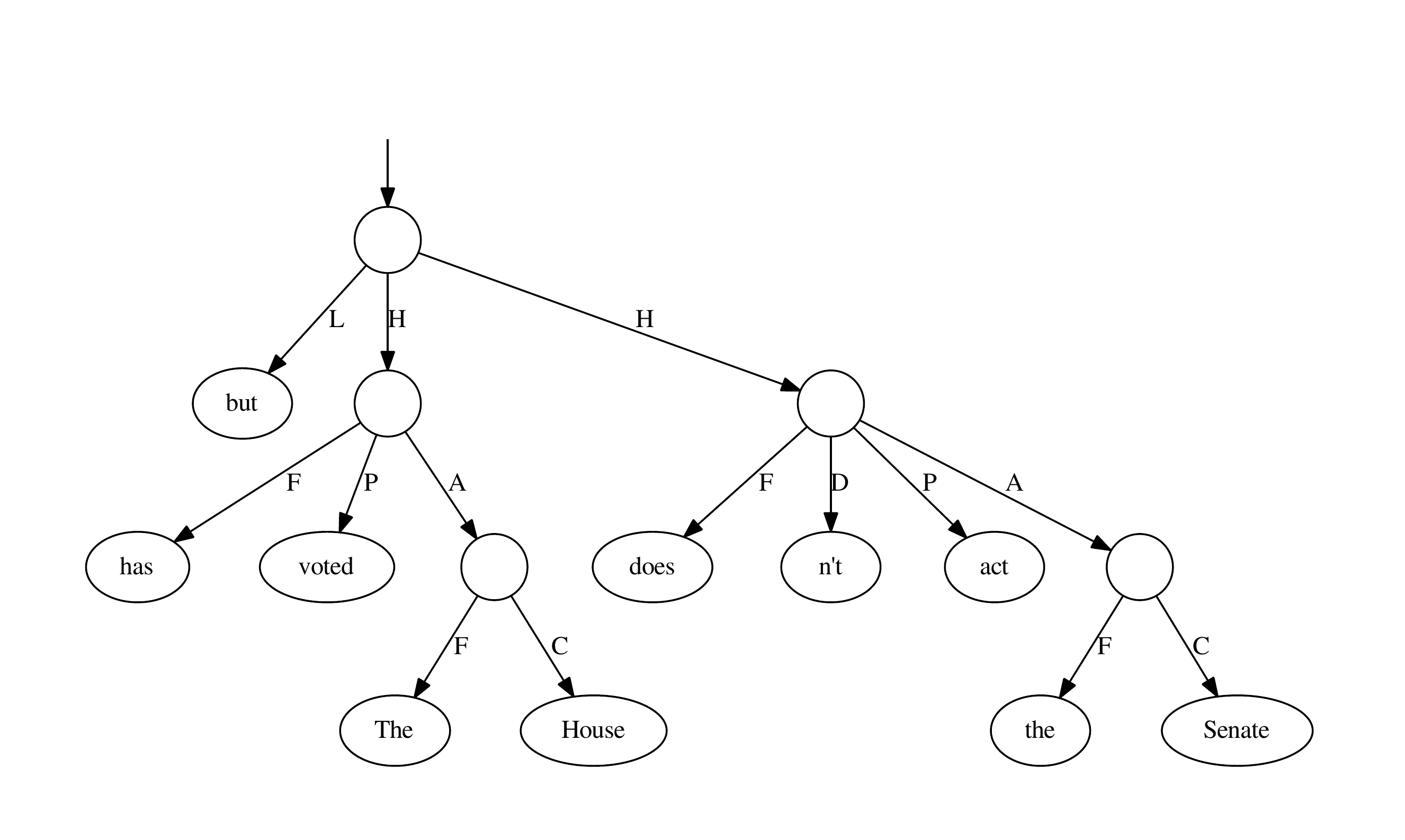}}
\caption{UCCA: Universal Conceptual Cognitive Annotation \citep{abend-rappoport-2013-ucca}}
\label{sfig:ucca}
\end{subfigure}
\hspace{5mm}
\begin{subfigure}{.51\textwidth}
\centering
\textsmaller[2]{\texttt{%
\begin{tabular}[t]{@{} l @{~} l @{\hspace{5mm}} l @{~} l@{}}
\toprule
\drgvar{b1} REF \drgvar{x1} &
    \comm{0:3}
& \drgvar{b3} REF \drgvar{x2} &
    \comm{24:27}
\\
\drgvar{b1} PRESUPPOSITION \drgvar{b2} &
    \comm{0:3}
& \drgvar{b3} PRESUPPOSITION \drgvar{b4} &
    \comm{24:27}
\\
\drgvar{b1} Name \drgvar{x1} "house" &
    \comm{4:9}
& \drgvar{b3} Name \drgvar{x2} "Senate" &
    \comm{28:34}
\\
\drgvar{b1} house "n.05" \drgvar{x1} &
    \comm{4:9}
& \drgvar{b3} senate "\posn{n}{01}" \drgvar{x2} &
    \comm{28:34}
\\
\drgvar{b2} REF \drgvar{e1} &
    \comm{14:19}
& \drgvar{b4} REF \drgvar{t2} & 
    \comm{35:39}
\\
\drgvar{b2} REF \drgvar{t1} &
    \comm{10:13}
& \drgvar{b4} EQU \drgvar{t2} "now" & 
    \comm{35:39}
\\
\drgvar{b2} TPR \drgvar{t1} "now" &
    \comm{10:13}
& \drgvar{b4} time "n.08" \drgvar{t2} &
    \comm{35:39}    
\\
\drgvar{b2} Time \drgvar{e1} \drgvar{t1} &
    \comm{10:13}
& \drgvar{b5} Time \drgvar{e2} \drgvar{t2} &
    \comm{35:39}
\\
\drgvar{b2} time "n.08" \drgvar{t1} &
    \comm{10:13}
& \drgvar{b4} NEGATION \drgvar{b5} &
    \comm{39:42}
\\
\drgvar{b2} Agent \drgvar{e1} \drgvar{x1} &
    \comm{14:19}
& \drgvar{b5} REF \drgvar{e2} &
    \comm{43:46}
\\
\drgvar{b2} vote "v.01" \drgvar{e1} &
    \comm{14:19}
& \drgvar{b5} Agent \drgvar{e2} \drgvar{x2} &
    \comm{43:46}
\\
\drgvar{b2} CONTRAST \drgvar{b4} &
    \comm{20:23}
& \drgvar{b5} act "v.01" \drgvar{e2} &
    \comm{43:46}
\\
\bottomrule
\end{tabular}
}}
\caption{DRS: Discourse Representation Structure in a clausal form \citep{kampreyle:drt,GMB:2017,abzianidze-etal-2017-parallel}}
\label{sfig:clf}
\end{subfigure}
\caption{The meaning representation graphs (a-d) of the MRP~2020 frameworks for the sentence \textit{The House has voted but the Senate doesn't act}.
(e) is the DRS of \autoref{fig:sample_drs} in the clausal form, a suitable format for semantic parsing.
The goal is to convert (e) into a graph somewhat similar to (a-d).}
\label{fig:contrast-mr}
\end{figure*}

Graphs are a common data structure for representing meaning of natural language sentences or texts.
Several shared tasks on semantic parsing have been organized, and the target meaning representations of the shared tasks were predominantly encoded as directed labeled graphs:%
\footnote{Throughout the paper, we mean a directed labeled graph when simply talking about graphs, unless stated otherwise.
}
Semantic Dependency Graphs \cite{oepen-etal-2014-semeval,oepen-etal-2015-semeval}, Abstract Meaning Representation \cite{may-2016-semeval,may-priyadarshi-2017-semeval}, 
and Universal Conceptual Cognitive Annotation \cite{hershcovich-etal-2019-semeval}.
Some of these graphs are presented in \autoref{fig:contrast-mr}.  
Recently, \citet{Oep:Abe:Haj:19} packaged several meaning representation graphs in a uniform graph abstraction and serialization for cross-framework meaning representation parsing.

Parallel to these developments our point of departure is Discourse Representation Theory (DRT, \citealp{kampreyle:drt}), a well-studied framework for studying formal semantics beyond sentences.
Its meaning representation structures, Discourse Representation Structure (DRS), are directly translatable into formal logic.
A sample DRS, in its traditional box format, is illustrated in \autoref{fig:sample_drs}.
We will discuss the DRS in more details in \autoref{s:drs}.   

Obviously, DRSs are meaning representation structures, but they are different from the already mentioned graph-based meaning representations in two aspects.
First, \textbf{DRSs are not inherently graphs}.
A DRS is more like a formula of predicate logic which is further organized in sub-formulas and governed with additional operations that account for co-reference and presupposition.
That's why DRSs are usually not considered as graph-based meaning representations.
For example, DRT was not among the frameworks of the shared task on cross-framework meaning representation parsing (MRP~2019, \citealp{Oep:Abe:Haj:19}) since the meaning representations at the shared task were all uniformly formatted as graphs.
\citet{zabokrtsky-et-al-2020} excluded DRSs when surveying sentence meaning representations as they ``limit [themselves] to meaning representations whose backbone structure can be described as a graph over words (possibly with added non-lexical nodes) [\ldots]''.
The second main contrast between DRSs and several of the 
graph-based meaning representations is that \textbf{DRSs are very different from syntactic structures}.
DRSs have roots in formal semantics, and they are geared to account for negation, quantification, and semantic scope rather than for syntactic structures.%
\footnote{For instance, this fact is another reason for excluding DRSs from the survey by \citet{zabokrtsky-et-al-2020}: ``we do not include primarily logical representations which are too distant from sentence structures; this leaves out some prominent frameworks such as the Groningen Meaning Bank [\ldots]''. 
}

Given that graphs are mainstream when it comes to representing meaning and semantically parsing wide-coverage natural language texts, it is important that DRSs are also convertible into graphs, and we refer to these structures as Discourse Representation Graphs (DRGs).
This will make DRSs accessible for researchers that primarily focus on graph-based meaning representations and parsing: (a) already existing graph-based semantic parsing models can be re-used or tested on DRGs; and (b) the specific structure of DRGs, reflecting formal semantics of the meaning, will pose new challenges for graph representation learning.

In a nutshell, to embrace DRSs in the second edition of the shared task on cross-framework (and cross-lingual) meaning representation parsing (MRP~2020; \citealp{Oep:Abe:Abz:20}), we investigate the conversion of DRSs from clausal form (the form adapted to semantic parsing, see \autoref{sfig:clf}) into graphs.
While doing so, our goal is to (i) make DRGs structurally as close as possible to the graphs of other frameworks in MRP~2020 (see \autoref{fig:contrast-mr}),
and (ii) keeping redundant information in DRGs to a minimum to prevent graphs of extensive size and to avoid inflation of the evaluation score.
Our efforts contribute to unified parsing models and evaluation tools across the frameworks.
Hopefully, it will also save the time of participants by preventing them from developing a completely new parsing model for DRGs.

\begin{figure*}[t!]
\begin{minipage}{.63\textwidth}
\captionsetup{type=table}
\scalebox{.93}{%
\begin{tabular}{@{}c @{~} c c l l@{}}
\toprule
Class & \multicolumn{2}{l}{Type symbol} & SDRS signature & Examples
\\\midrule[\heavyrulewidth]\addlinespace
\multirow{3}{*}{\rotatebox{90}{Entity}} & \multirow{2}{*}{$\bm{t}$} & \bmC{} & constant & $\sym{now}, \sym{house}, \sym{senate}$ 
\\
~ & ~ & $\bm{r}$ & discourse referent & $x_1, x_2, e_1, e_2, t_1, t_2$
\\\arrayrulecolor{gray}\cmidrule{2-5}\arrayrulecolor{black} 
~ & ~ & \bmB{} & box label & \bsty{b1}, \bsty{b2}, \bsty{b3}, \bsty{b4}, \bsty{b5}
\\\addlinespace\midrule[\heavyrulewidth]\addlinespace
\multirow{3}{*}{\rotatebox{90}{Predicate}} & \multirow{2}{*}{\bmB{}} & $\bm{S}$ & semantic role & $\sym{Agent}$, $\sym{Name}$, $\sym{Time}$
\\
~ & ~ & $\bm{M}$ & comparison relation & $\prec$, $=$
\\\arrayrulecolor{gray}\cmidrule{2-5}\arrayrulecolor{black}
~ & ~ & \bmC{} & concept & $\sym{house.n.05}$, $\sym{act.v.01}$
\\\addlinespace\midrule[\heavyrulewidth]\addlinespace
\multicolumn{2}{l}{\multirow{2}{*}{\tabul[l]{Discourse\\connective}}} & $\bm{R}$ & discourse relation & {\small $\sym{CONTRAST}$}
\\
~ & ~ & $\bm{O}$ & DRS operator & {\small $\sym{NEGATION}$, $\sym{PRESUPPOSITION}$}
\\\addlinespace\bottomrule
\end{tabular}
}
\caption{Classification of the DRS signature. Each element of the signature has a type symbol (in a bold font). $\bm{t}$ is for terms, which might be a constant or a discourse referent, while \bmB{} stands for binary relations, which are semantic roles and comparison relations.}
\label{tab:drs_signature}
\end{minipage}
\hspace{3mm}
\begin{minipage}{.33\textwidth}
\scalebox{.76}{
\fdrs{
  \begin{tabular}{@{}l@{~~}r@{}}
    \begin{tabular}[t]{@{}l}
    \pdrs{b1}{$x_1$}{
        $\sym{house.n.05}(x_1)$\\
        $\sym{Name}(x_1, \sym{house})$}
    \\[-1.3mm]
    \pdrs{b3}{$x_2$}{
        $\sym{senate.n.01}(x_2)$\hspace*{5mm}\\
        $\sym{Name}(x_2, \sym{senate})$}
    \\[-1.3mm]
    \pdrs{b4}{$t_2$}{
        $\sym{time.n.08}(t_2)$\\
        $t_2 = \sym{now}$}
    \end{tabular}
    &
    \begin{tabular}[t]{r@{}}
    \pdrs{b2}{$e_1$ ~ $t_1$}{
        $\sym{vote.v.01}(e_1)$\\
        $\sym{Agent}(e_1, x_1)$\\
        $\sym{Time}(e_1, t_1)$\\
        $\sym{time.n.08}(t_1)$\\
        $t_1 \prec \sym{now}$}
    \\[-2mm]
    \pdrs{b5}{$e_2$}{
        $\sym{act.v.01}(e_2)$\\
        $\sym{Agent}(e_2, x_2)$\\
        $\sym{Time}(e_2, t_2)$}
    \end{tabular}
  \end{tabular}
}{
    $\sym{PRESUPPOSITION}(\text{\bsty{b1,b2}})$\\
    $\sym{PRESUPPOSITION}(\text{\bsty{b3,b4}})$\\
    $\sym{CONTRAST}(\text{\bsty{b2,b4}})$\\
    $\sym{NEGATION}(\text{\bsty{b4,b5}})$
}}
\vspace{-1.3mm}
\caption{A flat visualization of a box-formatted DRS for the sentence \textit{The House has voted but the Senate doesn't act}.}
\label{fig:sample_drs}
\end{minipage}
\end{figure*}

The rest of the paper is organized as follows.
First, \autoref{s:drs} briefly describes the building blocks of DRSs, and then \autoref{s:existing} outlines already existing approaches of converting DRSs into graphs.
In addition to the existing ones, \autoref{s:new} introduces several candidate graph-based encodings of DRSs. 
In \autoref{s:matching}, we compare several DRG formats on the computational feasibility of finding maximum common edge subgraph (MCES) because the computational feasibility is crucial for evaluating the meaning representation graphs against the gold standard.
In the end, based on the findings of the MCES experiment and our desire for similarity with other graph-based frameworks, we select the specific DRG format that is included in MRP~2020.

\section{Discourse Representation Structures}\label{s:drs}

DRT is a framework that dates back to the early 1980s \citep{kamp81,Heim1982}.  
Since then, the framework has gone through several extensions and modifications to account for certain semantic or pragmatic phenomena. 
Throughout the paper we use DRSs that are derived from the Parallel Meaning Bank (PMB,  \citealp{abzianidze-etal-2017-parallel}). 
One such DRS is presented in \autoref{fig:sample_drs}.
The DRS signature is given in \autoref{tab:drs_signature}.

The PMB incorporates several extensions to DRSs.
On a micro level, the extensions aim to make DRSs language-neutral by disambiguating non-logical symbols with WordNet \cite{miller1995wordnet} synsets and VerbNet \cite{Bonial:11} roles, where the VerbNet roles are used in combination with neo-Davidsonian event semantics \cite{Parsons1990-PAREIT}.
On a macro level, presuppositions are modeled and explicitly represented following \citet{van_der_sandt:92} and Projective DRT \cite{venhuizen-etal-2013-parsimonious} while
discourse is analyzed following Segmented DRT \cite{asherlascarides} and flattened by treating discourse relations and DRS operators in a unified way.
Due to these extensions, all boxes are labeled with identifiers.

Let's decipher what the DRS in \autoref{fig:sample_drs} is expressing.
It consists of two parts: a set of boxes and a set of \emph{discourse connectives} applied to box labels (i.e., identifiers).
Boxes can be seen as sub-formulas whose separation is relevant for fine-grained semantics.
Each box includes a (possibly empty) set of \emph{discourse referents} stacked on a (possibly empty) set of \emph{conditions}.
The example sentence contains two clauses, corresponding to boxes \bsty{b2} and \bsty{b4}, that are related with each other via the $\sym{CONTRAST}$ discourse relation.  
Both \bsty{b2} and \bsty{b4} presuppose the existence of entities $x_1$ (for \textit{the House}) and $x_2$ (for \textit{the Senate}), which are further characterized with concepts (using WordNet synsets) and the naming semantic role.  
The presuppositions are put in separate boxes labeled with \bsty{b1} and \bsty{b3}.
The presupposition relations are explicitly stated with the binary $\sym{PRESUPPOSITION}$ DRS operator.
Since we use a flat visualization of DRSs, \bsty{b5}, which is negated and nested in \bsty{b4} (expressed by $\sym{NEGATION}(\text{\bsty{b4}}, \text{\bsty{b5}})$), is depicted outside \bsty{b4}.
In addition to modeling verb argument structure via neo-Davidsonian event semantics and semantic roles, the DRS also contains information about tense.%
\footnote{Note that $t_2$ is in \bsty{b4} because it has to be out of the scope of negation: there is a time $t_2$, and it is not the case that at $t_2$ \textit{the Senate acts}.
}


\begin{figure*}[t!]
\begin{subfigure}{.25\textwidth}\centering
\mbox{\includegraphics[clip, trim=248mm 39mm 14mm 4mm, 
    width=.84\textwidth]{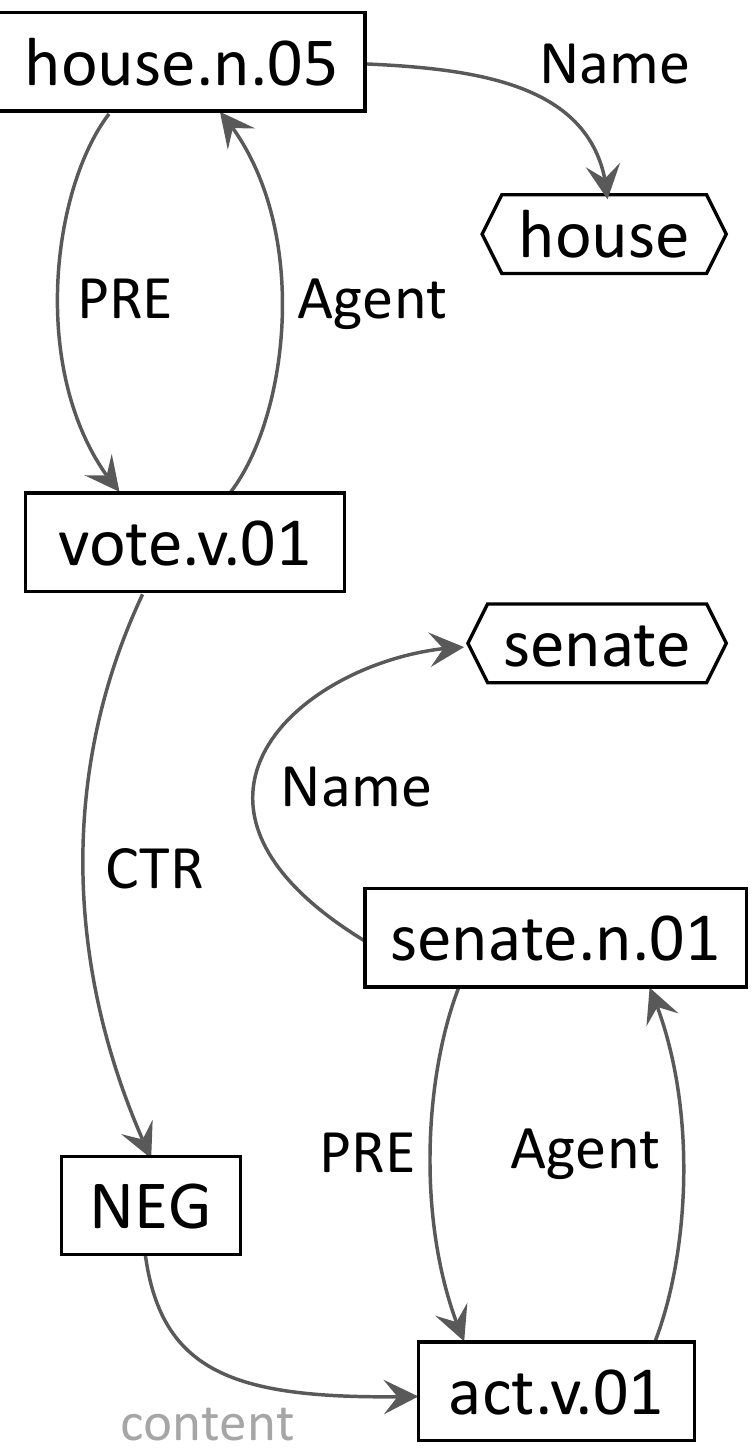}}
\caption{The augmented graph of \citet{Power99-enlgw} corresponding to the simplified sample DRS. 
The graph is a felicitous extension of \citeauthor{Power99-enlgw}'s original proposal over DRSs with presuppositions and discourse relations.}
\label{sfig:p99}
\end{subfigure}
\hspace{5mm}
\begin{subfigure}{.34\textwidth}\centering
\mbox{\includegraphics[clip, trim=0mm 36mm 219mm 1mm, 
    width=.93\textwidth]{src/drs/existing/BB-L18-P99.pdf}}
\caption{The \BBenc{} encoding largely follows \citet{basile-bos-2013-aligning} and incorporates several additional simplifications. 
The encoding is node-centric.
\bmB{} and \bmC{} are encoded as labeled nodes while $\bm{R}$, $\bm{O}$ and argument positions ($A$) as labeled edges. 
Only \bmB{} and $\bm{r}$ are unlabeled nodes.}
\label{sfig:bb_enc}
\end{subfigure}
\hspace{5mm}
\begin{subfigure}{.33\textwidth}\centering
\mbox{\includegraphics[clip, trim=132mm 36.5mm 101mm 1.5mm, 
    width=.85\textwidth]{src/drs/existing/BB-L18-P99.pdf}}
\caption{\Liu{$18$} is the edge-centric encoding by \citet{liu-etal-2018-discourse}. 
\bmB{} and \bmC{} are represented as unlabeled nodes with \bmB{}- and \bmC{}-labeled incoming edges.
$\bm{R}$, $\bm{O}$ and argument positions ($A$) are encoded as labeled edges.
Unlabeled nodes are introduced not only by \bmB{} and $\bm{r}$ but also by \bmB{} and \bmC{}.}
\label{sfig:l18}
\end{subfigure}
\caption{Contrasting the existing graph representations of DRSs. The graphs encode the sample DRS from \autoref{fig:sample_drs}. For brevity, the tense information is omitted from the DRS. Unlabeled nodes have a gray background. The shapes of nodes are not part of the graphs but simply help with reading to distinguish the types of symbols.}
\label{fig:existing-drgs}
\end{figure*}

\section{Related Work}\label{s:existing}

There have been several approaches to represent DRSs as graphs.
These representations are put side-by-side in \autoref{fig:existing-drgs}.

The work by \citet{Power99-enlgw} doesn't aim to convert DRSs into graphs as such, but it proposes to augment object-oriented knowledge representation (OOKR) graphs with additional scope information to establish correspondence with DRSs.
Although the correspondence is incomplete, e.g., some OOKR graphs might have no corresponding DRS.
The augmentation of \citet{Power99-enlgw} doesn't cover DRSs with discourse relations, presuppositions (e.g., \texttt{b1} to \texttt{b2} in \autoref{fig:sample_drs}) or with an embedded box that contains base and complex conditions (like \texttt{b4} in \autoref{fig:sample_drs}).
Nevertheless, for demonstration purposes, we still present \citet{Power99-enlgw}'s augmented graph for a felicitous, simplified DRS of \autoref{fig:sample_drs}.

\citet{basile-bos-2013-aligning} proposed converting DRSs into graphs, calling them Discourse Representation Graphs (DRGs).
Their goal was to facilitate word-level alignment between surface forms and the corresponding DRSs to generate texts from DRSs.
The graph encoding, with several simplifications, is exemplified in \autoref{sfig:bb_enc}.%
\footnote{Originally \citet{basile-bos-2013-aligning} use more labels for edges that expresses type-specific information of nodes.
For example, they use different edge labels to distinguish the first argument position of \bmB{} from the only argument position of \bmC{} while in the paper we use the same label for both.
\citet{basile-bos-2013-aligning} also encodes $\bm{O}$ as a reified node that introduces two edges $\texttt{b4}\xrightarrow{\text{unary}}\text{NEG}\xrightarrow{\text{scope}}\texttt{b5}$.
Instead, we simply model $\bm{O}$ with a single edge $\texttt{b4}\xrightarrow{\text{NEG}}\texttt{b5}$.
\label{fn:bb}
}
The simplifications decrease the number of nodes and out-of-signature labels in the graph.
The encoding can be seen as node-centric since the most frequent signature symbols, namely the symbols of type \bmB{} and \bmC{}, are modeled as labeled  nodes.
Argument positions ($A$) of binary predicates are distinguished via edge labels.
We call this DRG format \BBenc{}.

To evaluate the output of their DRS parser, \citet{liu-etal-2018-discourse} converted DRSs into graphs, demonstrated in \autoref{sfig:l18}.
This graph encoding, in contrast to \BBenc{}, is edge-centric as the symbols of type \bmB{} and \bmC{} are used as edge labels.
Moreover, compared to \BBenc{}, the encoding contains more unlabeled nodes since \bmB{} and \bmC{} are also modeled with reified nodes.
We call \citet{liu-etal-2018-discourse}'s encoding \Liu{$18$}. 

Interestingly, in contrast to the proposed graph encodings of DRS, \citet{van-noord-etal-2018-evaluating} refused to convert DRSs into graphs and instead used so-called \emph{clausal form} of DRSs (see \autoref{sfig:clf}).
The clauses in clausal form are triples, e.g., \mbox{$\langle$\bsty{b4}, $\sym{NEGATION}$, \bsty{b5}$\rangle$}, or quadruples, e.g., \mbox{$\langle$\bsty{b2}, $\sym{Agent}$, \bsty{e1}, \bsty{x1}$\rangle$}, where the quadruples are hyper-edges and fall out of the scope of standard graph encodings.
The official evaluation of the shared task on DRS parsing \cite{abzianidze-etal-2019-first} was also based on clausal form of DRSs.

\section{More Graph-based Encodings of DRS}\label{s:new}

As illustrated in the previous section, there is no agreement on how DRSs should be converted into graphs (or whether they should be converted at all). 
The range of graph encodings in \autoref{fig:existing-drgs} presents anything but an exhaustive list.
Some encoding can even be further refined and compressed without affecting the readability or expressivness.
For instance, as explained in \autoref{fn:bb}, \BBenc{} represents a refined version of DRGs proposed by \citet{basile-bos-2013-aligning}.
\Liu{$18$} can also be further compressed by discarding reified concept nodes and their outgoing \graytxt{a1} edges, e.g., replacing \bsty{b5}$\xrightarrow{\sym{act.v.01}}$\bsty{c8}$\xrightarrow{\text{\graytxt{a1}}}e_2$ with \bsty{b5}$\xrightarrow{\sym{act.v.01}}e_2$.
We will use \Liu{$18^*$} to refer to the DRGs refined in such a way.

In general, the choices in which DRG formats might differ are several.
Here we will discuss some of them, namely (see also \autoref{tab:encoding_choices}):

\begin{enumerate}[label={(\Alph*)}]\itemsep-1mm\topsep-3mm
    \item Expressing $A$rgument positions of \bmB{} via forking and labeled edges (\rolfork{\bmB{}}\,, like \BBenc) or solely via graph configuration (\rolmid{\bmB{}}\,, without labeled edges), e.g., encoding $\sym{Agent}(e_1, x_1)$ as ${e_1\xrightarrow{}\sym{Agent}\xrightarrow{}x_1}$; 
    \label{it:A}
    
    \item Representing \bmB{}inary predicates as labeled nodes (\gnode{\bmB{}}, like \BBenc) or unlabeled nodes with \bmB{}-labeled edges (\toreif{\bmB{}}\,, like \Liu{$18$});
    
    \item Encoding \bmC{}oncepts as labeled nodes (\gnode{\bmC{}}, like \BBenc), unlabeled nodes with incoming \bmC{}-labeled edges (\toreif{\bmC{}}\,, like \Liu{$18$}), labeled edges (\edgetoref{\bmC{}}\,, like \Liu{$18^*$}), or as a label on an $\bm{r}$ node (\cnodeonref{\bmC{}}\,, which will be discussed further);
    
    \item[(I)] Expressing box membership explicitly (Exp) or implicitly (Imp).
    Whether a node (corresponding to \bmB{}, \bmC{}, or $\bm{r}$) is \emph{In} a particular \bmB{}, can be depicted via an explicit connecting edge or implicitly via graph configuration. 
\end{enumerate}

\begin{table}[t]
\scalebox{1}{
\begin{tabular}{@{}l @ {~~} c c @{~~~} c c@{}}
\toprule
DRG encoding & $A$rgs & \bmB{} & \bmC{} & \emph{I}n-box 
\\
\midrule[\heavyrulewidth]\addlinespace
\BBenc{}\footnotesize{+typed edges}  & \rolfork{\bmB{}} & \gnode{\bmB{}} & \gnode{\bmC{}} & Exp
\\
$\fargs\Brln\Crln$   & \rolfork{\bmB{}} & \gnode{\bmB{}} & \gnode{\bmC{}} & Exp
\\
$\fargs\Ctorn\Btorn$ (\Liu{$18$})  & \rolfork{\bmB{}} & \toreif{\bmB{}} & \toreif{\bmC{}} & Exp
\\
$\fargs\Btorn\Ctor$ (\Liu{$18^*$})  & \rolfork{\bmB{}} & \toreif{\bmB{}} & \edgetoref{\bmC{}} & Exp
\\
\arrayrulecolor{gray}\midrule\arrayrulecolor{black}
$\fargs\Brln\Ctor$  & \rolfork{\bmB{}} & \gnode{\bmB{}} & \edgetoref{\bmC{}} & Exp
\\
$\bargs\Brln\Ctor$ & \rolmid{\bmB{}} & \gnode{\bmB{}} & \edgetoref{\bmC{}} & Exp 
\\
\arrayrulecolor{gray}\midrule\arrayrulecolor{black}
$\fargs\Btorn\Cor$ & \rolfork{\bmB{}} & \toreif{\bmB{}} & \cnodeonref{\bmC{}} & Exp 
\\
$\fargs\Brln\Cor$ & \rolfork{\bmB{}} & \gnode{\bmB{}} & \cnodeonref{\bmC{}} & Exp 
\\
$\bargs\Brln\Cor$ & \rolmid{\bmB{}} & \gnode{\bmB{}} & \cnodeonref{\bmC{}} & Exp 
\\
\arrayrulecolor{gray}\midrule\arrayrulecolor{black}
$\bargs\Brln\Cor I$ & \rolmid{\bmB{}} & \gnode{\bmB{}} & \cnodeonref{\bmC{}} & Im-a1 
\\
$\bargs_a\Brln\Cor I$ & \rolmid[1.4pt]{\bmB{}} & \gnode{\bmB{}} & \cnodeonref{\bmC{}} & Im-a1 
\\
\bottomrule
\end{tabular}}
\caption{Several combinations of the choices in DRG design.
The choices concern representation of argument positions, \bmB{} symbols, \bmC{} symbols, and in-box relations.
The names of encodings visually follow the combinations of the choices.
}
\label{tab:encoding_choices}
\end{table}

Here we would like to elaborate more on (I).
The box membership in DRT directly accounts for a semantic scope.
Like discourse referents, conditions are also members of boxes.
So, we also need to express the box membership of condition predicates in the graphs.
All the encodings in \autoref{fig:existing-drgs} explicitly express box membership.
For instance, $\sym{Agent}(e_1,x_1)$ belonging to \bsty{b2} is expressed via connecting \bsty{b2} to the \sym{Agent} node (see \autoref{sfig:bb_enc}) or via the outgoing \sym{Agent} edge from \bsty{b2} to \bsty{c3}. 
Explicating all box memberships via labeled edges increases the graphs in size.
To prevent this, one can make box membership of certain predicates or their arguments implicit but at the same time easily and unambiguously recoverable from the graphs.
For example, if we assume that directionality of arrows carries the in-box inheritance and consider the case when argument positions are configurationally encoded (\rolmid{\bmB{}}),
then there is no need to explicate the in-box relation for $\sym{Name}$ in ${x_1\xrightarrow{}\sym{Name}\xrightarrow{}\sym{house}}$ whenever the $\sym{Name}$ condition and $x_1$ are in the same box.%
\footnote{Remember that a discourse $\bm{r}$eferent is considered to be \emph{in a box} if it is introduced in the top row of the box.
}
We dub such an implication of box membership of \bmB{} from the first argument as `Im-a1'.

\autoref{tab:encoding_choices} lists several DRG formats based on combinations of how argument positions, binary predicates, concepts, and in-box relations are represented in a graph.
While modeling the argument position, \rolmid{\bmB{}} is preferred over \rolfork{\bmB{}} from a theoretical point of view because a$1$ and a$2$ labels are not part of the DRS signature;
They are ad-hoc ingredients only helping with distinguishing argument positions.
When it comes to modeling concepts, as we already discussed, \edgetoref{\bmC{}} leads to more economic graphs than \toreif{\bmC{}}.


\begin{figure}
\centering
\mbox{\includegraphics[clip, trim=9mm 41mm 204mm 4mm, 
    width=.4\textwidth]{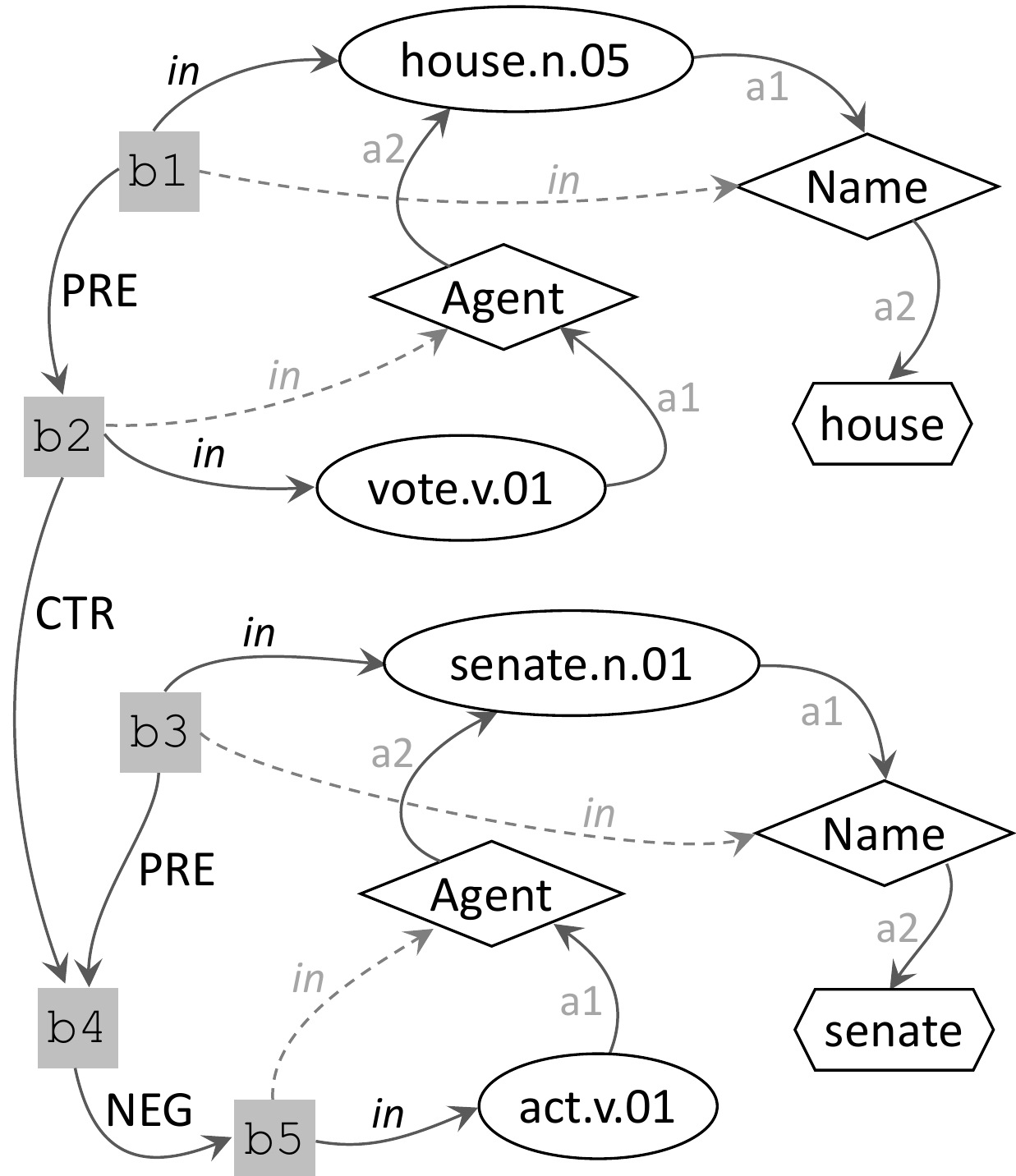}}
\caption{$\bargs_a\Brln\Cor$ and $\bargs\Brln\Cor I$ encodings. 
The $\bm{r}$ nodes are labeled with \bmC{}oncepts and connected to the boxes via \emph{in}-edges.
Dashed \emph{in}-edges for \bmB{}inary predicates and the a$N$ edge labels are recoverable.
$\bargs\Brln\Cor I$ is obtained by ignoring dashed edges and gray edge labels.
Unlabeled nodes are colored in grey where their labels merely serve to match graph components across the different visualizations.}
\label{fig:new_DRG}
\end{figure}

In the PMB annotation, for almost any discourse referent, there exists the most specific concept among the concepts applied to it.
For example, a discourse referent might have only two concepts, $\sym{male.n.02}$ and $\sym{person.n.01}$, applied to it, but among these concepts there exists the most specific concept, namely  $\sym{male.n.02}$, as $\sym{male.n.02}$ is a hyponym of $\sym{person.n.01}$ according to WordNet. 
The \cnodeonref{\bmC{}} choice exploits this annotation property of concepts in the PMB and labels the node of a discourse referent with the corresponding most specific concept.
This type of encoding of \bmC{} is shown in \autoref{fig:new_DRG}. 

\autoref{fig:new_DRG} also depicts $\bargs\Brln\Cor I$ DRG encoding with implicit box membership of \bmB{}.
Though all the box membership edges of \bmB{} are made implicit in the encoding example, this is not the case in general.
For example, attributive and predicative adjectives usually introduce \mbox{$\langle$\bsty{b1}, $\sym{Attribute}$, \bsty{x1}, \bsty{s1}$\rangle$} clause, where \bsty{x1} is the attributed entity which is not necessarily introduced in the same \bsty{b1} box as the attributing state \bsty{s1}.
Another example is a construction with a locative preposition and a definite noun phrase, e.g., \emph{hid a parcel under the bed}, whose DRS contains the following fragment:
\begin{center}
\scalebox{.8}{\texttt{%
\begin{tabular}[t]{l l}
\drgvar{b2} REF \drgvar{e1} & \drgvar{b2} Location \drgvar{e1} \drgvar{x3}\\
\drgvar{b2} hide "v.01" \drgvar{e1} & \drgvar{b2} SZP \drgvar{x2} \drgvar{x3}\\ 
\drgvar{b2} REF \drgvar{x1} & \drgvar{b3} REF \drgvar{x2}\\
\drgvar{b2} parcel "n.01" \drgvar{x1} & \drgvar{b3} bed "n.01" \drgvar{x2}\\
\drgvar{b2} Patient \drgvar{e1} \drgvar{x1}\\
\end{tabular}}}
\end{center}
where the binary relation $\sym{SZP}$ (spatial above) is in a different box than its first argument. 

As we have shown, there are at least a dozen ways to dress up DRSs as graphs.
Some of the DRG formats are verbose, some can employ default rules to ignore certain redundancies, some require out-of-signature symbols, and some prefer labeled edges over labeled nodes.
There isn't enough space to illustrate the graphs listed in \autoref{tab:encoding_choices}, but each of the mentioned encoding choices is demonstrated by at least one of the graphs from \autoref{fig:existing-drgs} and \autoref{fig:new_DRG}.

\section{Matching \& Evaluating DRGs}\label{s:matching}

In graph-based semantic parsing, system outputs are conventionally evaluated against the gold standard graphs by finding the maximum common edge subgraph (MCES) for each pair of produced and gold graphs, and then calculating macro-average F-score \citep{Oep:Abe:Haj:19}.
In general, the MCES problem is NP-complete, and finding the maximum subgraph \emph{shared} between two relatively large graphs is sometimes computationally infeasible.
In this section, we experiment on how computationally expensive is the MCES problem for each DRG design.

\begin{table}[t]
\hspace{-2mm}
\begin{tabular}{@{} l @{~} r r @{~~~} r @{~~~} r @{~~~} r @{~~~} r @{}}
\multirow{2}{*}{\parbox{10mm}{DRS\\parsers}}  
            & 
            & \rotatebox{75}{\BBenc}
            & \rotatebox{75}{$\fargs_a\Btorn\Ctorn$}
            & \rotatebox{75}{$\fargs_a\Brln\Ctor$}
            & \rotatebox{75}{$\fargs_a\Brln\Cor$}
            & \rotatebox{75}{$\bargs\Brln\Cor I$}
\\\cmidrule{3-7}
 & DRS & \multicolumn{5}{c}{DRGs}
\\\toprule
\poorLSTM{} & 64.6 & 79.6 & 74.3 & 77.7 & 77.9 & 78.2
\\\midrule
\Boxerdw{}  & 78.2 & 89.5 & 86.8 & 87.5 & 87.6 & 87.7
\\\midrule
\tacl{}     & 84.3 & 92.3 & 88.4 & 90.9 & 90.9 & 91.1
\\\midrule
\Boxerup{}  & 87.2 & 94.2 & 92.3 & 92.9 & 92.9 & 93.0
\\
\bottomrule
\end{tabular}
\caption{Macro F-scores of the models when their output is treated as DRS or DRG.
F-score for DRS is computed with \texttt{Counter} while for DRG with \texttt{mtool}. 
}
\label{tab:f_scores}
\end{table}

\begin{table*}[t]
\begin{tikzpicture}[baseline=(1.base), node distance=9.2mm]
\tikzset{every node/.style={anchor=west,text width=4em}}
\node (0) {};
\node (1) [right of=0,xshift=27mm,rotate=65] {\BBenc};
\node (2) [right of=1,rotate=65] {$\fargs\Brln\Crln$};
\node (3) [right of=2,rotate=65] {$\fargs\Ctorn\Btorn$};
\node (4) [right of=3,rotate=65] {$\fargs\Btorn\Ctor$};
\node (5) [right of=4,rotate=65] {$\fargs\Brln\Ctor$};
\node (6) [right of=5,rotate=65] {$\bargs\Brln\Ctor$};
\node (65) [right of=6,rotate=65] {$\bargs_a\Brln\Ctor$};
\node (7) [right of=65,rotate=65] {$\fargs\Btorn\Cor$};
\node (8) [right of=7,rotate=65] {$\fargs\Brln\Cor$};
\node (9) [right of=8,rotate=65] {$\bargs\Brln\Cor$};
\node (10)[right of=9,rotate=65] {$\bargs_a\Brln\Cor$};
\node (11)[right of=10,rotate=65] {$\bargs\Brln\Cor I$};
\node (12)[right of=11,rotate=65] {$\bargs_a\Brln\Cor I$};
\end{tikzpicture}
\\[-7mm]
{\renewcommand{\arraystretch}{1.09}
\begin{tabular}[b]{@{}l @{~~~} r r r}
DRS parser &\ding{56} &\CorCut &  
\\\midrule
\poorLSTM{} & 13 & 11& 
\\\midrule 
\Boxerdw{}  & 7 & 9&
\\\midrule
\tacl{}     & 4 & 3&
\\\midrule
\Boxerup{}  & 7 & 3&
\\\midrule
\end{tabular}}%
\kern-2mm
\raisebox{1pt}{\mbox{\includegraphics[clip, trim=58mm 45mm 30mm 8.5mm, width=.75\textwidth]{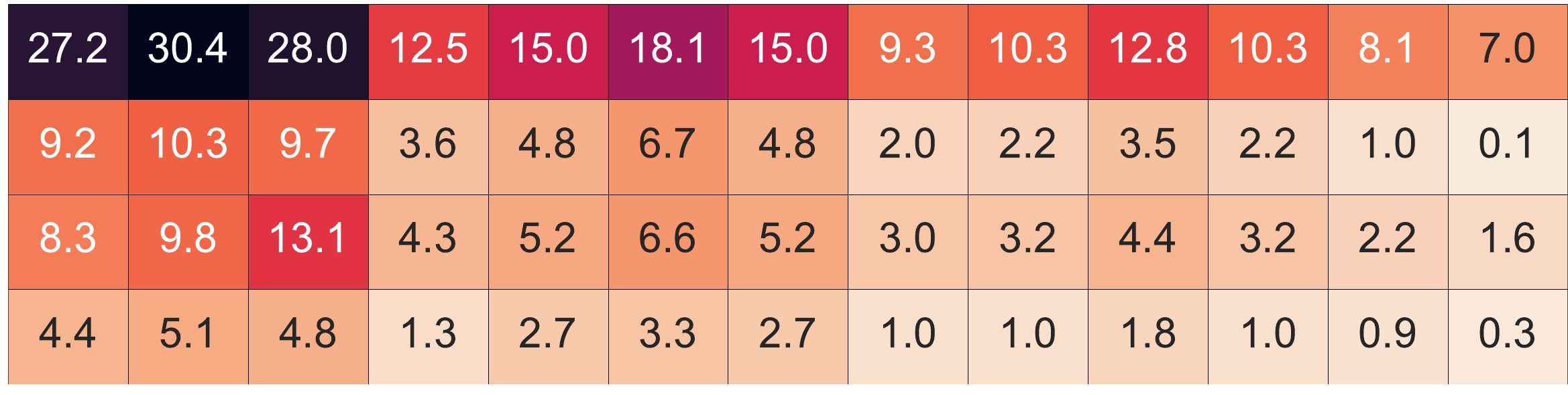}}}%
\caption{The percentage of approximate (i.e., non-exact) matches w.r.t.\ the total non-null DRGs.
Lower numbers are better as more graph matches corresponding to MCES are found.  
The total number of DRSs is 885. While converting DRSs into DRGs, \ding{56}-number of DRGs become null due to ill-formed DRSs and are excluded during calculating the percentages.
Encoding with \Cor{} additionally renders \CorCut-number of DRSs untranslatable.}
\label{tab:non_exact}
\end{table*}

\subsection{Data \& Tools}\label{ss:data}

We run the experiments on the output of existing DRS parsers.
Four distinct parsing models are selected to achieve diversity in the system output graphs.
Two of the parsers are end-to-end character-based LSTM models from \citet{van-noord-etal-2018-exploring}: one is their best model (\tacl) while another one is trained on fewer data on purpose to have mediocre performance (\poorLSTM).
Another two parsers are based on the semantic parser Boxer \citep{step2008:boxer}, which is used in the PMB to \emph{pack} all annotations layers into DRS boxes.
\Boxerdw{} is Boxer based on the NLP tools of the PMB pipeline\footnote{\url{https://pmb.let.rug.nl/software.php}}, on the other hand, \Boxerup{} is Boxer employing annotation layers output by \textsc{MaChAmp} \citep{vandergoot-etal-2020-machamp}.
As the names suggest, \Boxerup{} is a better model than \Boxerdw{}.  
The output DRSs are obtained by parsing the development set (885 documents) of the PMB v3.0.0.\footnote{\url{https://pmb.let.rug.nl/data.php}}
Evaluation of the models based on the DRSs of the dev set is given in \autoref{tab:f_scores}.
DRSs are scored with \texttt{Counter} \cite{van-noord-etal-2018-evaluating}, the clause matching tool for DRSs in clausal form.%
\footnote{\url{https://github.com/RikVN/DRS_parsing}}

For MCES-based matching of DRGs, we use \texttt{mtool}\footnote{\url{https://github.com/cfmrp/mtool}}, the Swiss Army Knife for Graph-Based Meaning Representation.
Based on the graph configurations, \texttt{mtool} schedules potential node-to-node mappings between two graphs.
This information is used to initialize \emph{promising} node-to-node mappings that might lead to finding the MCES early.  
\texttt{mtool} is the official scorer in both the MRP~2019 and MRP~2020 shared tasks.

All types of graph encodings employed in the experiments are obtained with the \texttt{DRS2Graph} tool.\footnote{\url{https://github.com/kovvalsky/DRS2Graph}}
This new converter from clause-based DRSs to labeled directed graphs is one of the contributions of the paper.

\subsection{Results \& Analysis}\label{ss:results}

The results of finding MCES between the system generated and converted DRGs and reference DRGs are provided in \autoref{tab:non_exact}.
The reference DRGs were obtained by converting the gold standard DRS of the PMB 3.0.0 development set.
We run experiments with 13 DRG formats.
All 885 DRSs were converted in each DRG format without problems.
In principle, the encodings with the \Cor{} choice are lossy, however, they were successfully applied to the gold DRSs.
Several parser-produced DRSs were not converted according to the \Cor{} choice since the parsers assert the inconsistent concepts for discourse referents.
For example, \Boxerup{} produced a DRS with $\sym{measure.n.02}$ and $\sym{book.n.01}$ applied to the same discourse referent.
Since these senses are not in hyponymy/hypernymy relation, the DRS didn't meet the requirement from \Cor{} and was one of the three DRSs of \Boxerup{} that couldn't be dressed up as \Cor{}-based graphs.%
\footnote{From 6620 gold DRSs of the PMB 3.0.0 training part, only 16 (0.24\%) DRSs didn't satisfy the constraint of \Cor{}.
}

\autoref{tab:non_exact} shows the computational (in)feasibility of the MCES problem across the combinations of parsing models and graph encodings (using the \texttt{mtool} implementation with default limits on its search space).
Given that models are sorted according to their performance in ascending order from top to bottom, the table shows that for relatively distinct graphs it can be difficult to guarantee the MCES solution.%
\footnote{When exhausting its search space limits, \texttt{mtool} falls back to an anytime strategy, returning the best overall match found up to that point.  This match will often correspond to the MCES, but there is no correctness guarantee in this mode.}
But things are not so straightforward as \tacl{} outperforms \Boxerdw{} but finding MCES for \Boxerdw{} is easier for 10 encodings out of 13.
This can be explained by the fact that gold DRSs are obtained from \Boxerdw{} while taking into account added human annotations.
Given this, it is expected that gold and \Boxerdw{}'s DRSs have in common substantial chunks of boxes, and this sharing is transferred on the DRGs too.   

Interestingly, the encodings \BBenc{} \citep{basile-bos-2013-aligning} and $\fargs_a\Btorn\Ctorn$ \citep{liu-etal-2018-discourse} are one of the most inefficient encodings across all the models.
For instance, non-exact (i.e., approximate) MCES was found for 237 DRG pairs out of 872 for \poorLSTM{} and  \BBenc{} encoding. 
For other encodings the ratio of approximate matches halves. 

Among the encodings with the $\Ctor$ choice, $\fargs_a\Btorn\Ctor$ appears to provide most computationally \emph{friendly} graphs.
Every encoding with $\Ctor$ becomes even better when $\Ctor$ is replaced with $\Cor$.
This is because $\Cor$ brings at least a 16\% reduction in the number of edges and increases the number of labeled nodes.
The latter apparently helps \texttt{mtool} to get better initializations for node mappings.

$\fargs_a\Btorn\Cor$ is the best among $\Cor$-featured encodings with explicit box membership.
It doesn't improve further when changing its encoding choices, including switching to $\bargs$.
The results show that $\fargs_a$ is consistently better than $\bargs$.
Even when they are combined, $\bargs$ adds no value to $\fargs_a$.
However, the advantage of $\bargs$ over $\fargs_a$ is that it configurationally distinguishes argument positions and there is no need for out-of-signature labels.
Moreover, $\bargs$ invites the intuitive inheritance property about in-box relation (see (I) discussed in \autoref{s:new}).
When incorporating the implicit in-box relation with $\bargs$, the combination $\bargs\Brln\Cor I$ yields a substantial decrease in the number of approximate matches.
This is explained by the fact that the number of edges decreases by at least 23\%.
Adding the out-of-signature edge labels for marking argument positions further improves the encoding. 

Differences between F-scores calculated over DRS (with \texttt{Counter}) and DRGs (with \texttt{mtool}) are significant (see \autoref{tab:f_scores}).
The gap between low- and high-performing model is greater than 10\% and 5\%, respectively.
The DRS-based score is more strict than the DRG-based one because DRSs are evaluated in the clausal form, where some DRSs conditions (e.g., built with $\bmB$) are modeled via quadruples, i.e., hyper-edges.
In DRGs, the hyper-edges are represented by multiple triples ($\langle \texttt{nodeID}, \texttt{edgeLabel}, \texttt{nodeID} \rangle$ or $\langle \texttt{nodeID}, \text{label}, \texttt{labelValue} \rangle$), and this additionally rewards the models when they get parts of hyper-edges correctly.  

\section{Conclusion}\label{s:conclusion}

There have been several approaches that encoded DRSs as graphs (surveyed in \autoref{s:existing}), but their objectives were to transform DRSs in a suitable format for particular applications rather than exploring and comparing different types of DRG encodings.
This paper fills this gap.
We have systematically characterized a dozen of DRG encodings and contrasted them with each other, and compared them to the DRS clausal form from an evaluation perspective.

We opt for the $\bargs\Brln\Cor I$ DRG encoding (see \autoref{fig:new_DRG}) to represent DRSs at the MRP~2020 shared task.
Despite the encoding being lossy, it represents an excellent trade-off due to the advantages it brings:
(a) the encoding has at least 23\% fewer edges than other encodings, which makes the DRGs more compact and easier to read;
(b) given that scope information inflates DRSs, learning relatively compact DRGs seems a good starting point for the shared task;
(c) only less than 0.25\% DRSs are lost when applying the encoding;
(d) it doesn't employ the out-of-signature labels \sym{a1} and \sym{a2};
(e) for the DRGs obtained from the average-performing DRS parsers, the evaluation tool can find exact maximal matches for at least 98.4\% of DRG pairs. 

When abstracting from the reification of the roles as nodes, the chosen DRG encoding and the graphs of other frameworks in MRP~2020 have abstractly parallel graph topologies for linguistically parallel predicate-argument structures.

\section*{Acknowledgments}

We thank Rik van Noord for providing us with outputs of the DRS parsers.
We acknowledge access to the Peregrine HPC cluster provided by the CIT of the University of Groningen,
and to the NLPL infrastructure provided by Sigma2 in Norway.
The first two authors were supported by the NWO-VICI grant (288-89-003).
The first author was additionally supported by the European Research Council (ERC) under the European Unions Horizon 2020 research and innovation programme (grant agreement No.\,742204).

\newpage

\bibliographystyle{acl_natbib}
\bibliography{mrp,refs}

\providecommand{\fromto}[2]{#1$\,$--$\,$#2}\providecommand{\pages}[3]{#2$\,$--$\,$#3}\providecommand\Beek[1]{\mbox{#1}}\providecommand\Noord[1]{\mbox{#1}}\providecommand\Lohuizen[1]{\mbox{#1}}
\begin{thebibliography}{33}
\expandafter\ifx\csname natexlab\endcsname\relax\def\natexlab#1{#1}\fi

\bibitem[{Abend and Rappoport(2013)}]{abend-rappoport-2013-ucca}
Omri Abend and Ari Rappoport. 2013.
\newblock \href {https://www.aclweb.org/anthology/W13-0101} {{UCCA}: A
  semantics-based grammatical annotation scheme}.
\newblock In \emph{Proceedings of the 10th International Conference on
  Computational Semantics ({IWCS} 2013) {--} Long Papers}, pages 1--12,
  Potsdam, Germany. Association for Computational Linguistics.

\bibitem[{Abzianidze et~al.(2017)Abzianidze, Bjerva, Evang, Haagsma, van Noord,
  Ludmann, Nguyen, and Bos}]{abzianidze-etal-2017-parallel}
Lasha Abzianidze, Johannes Bjerva, Kilian Evang, Hessel Haagsma, Rik van Noord,
  Pierre Ludmann, Duc-Duy Nguyen, and Johan Bos. 2017.
\newblock \href {https://www.aclweb.org/anthology/E17-2039} {The {P}arallel
  {M}eaning {B}ank: Towards a multilingual corpus of translations annotated
  with compositional meaning representations}.
\newblock In \emph{Proceedings of the 15th Conference of the {E}uropean Chapter
  of the Association for Computational Linguistics: Volume 2, Short Papers},
  pages 242--247, Valencia, Spain. Association for Computational Linguistics.

\bibitem[{Abzianidze et~al.(2019)Abzianidze, van Noord, Haagsma, and
  Bos}]{abzianidze-etal-2019-first}
Lasha Abzianidze, Rik van Noord, Hessel Haagsma, and Johan Bos. 2019.
\newblock \href {https://www.aclweb.org/anthology/W19-1201} {The first shared
  task on discourse representation structure parsing}.
\newblock In \emph{Proceedings of the {IWCS} Shared Task on Semantic Parsing},
  Gothenburg, Sweden. Association for Computational Linguistics.

\bibitem[{Asher and Lascarides(2003)}]{asherlascarides}
N.~Asher and A.~Lascarides. 2003.
\newblock \href {http://books.google.com.au/books?id=VD-8yisFhBwC}
  {\emph{Logics of conversation}}.
\newblock Studies in natural language processing. Cambridge University Press.

\bibitem[{Banarescu et~al.(2013)Banarescu, Bonial, Cai, Georgescu, Griffitt,
  Hermjakob, Knight, Koehn, Palmer, and
  Schneider}]{banarescu-etal-2013-abstract}
Laura Banarescu, Claire Bonial, Shu Cai, Madalina Georgescu, Kira Griffitt, Ulf
  Hermjakob, Kevin Knight, Philipp Koehn, Martha Palmer, and Nathan Schneider.
  2013.
\newblock \href {https://www.aclweb.org/anthology/W13-2322} {Abstract {M}eaning
  {R}epresentation for sembanking}.
\newblock In \emph{Proceedings of the 7th Linguistic Annotation Workshop and
  Interoperability with Discourse}, pages 178--186, Sofia, Bulgaria.
  Association for Computational Linguistics.

\bibitem[{Basile and Bos(2013)}]{basile-bos-2013-aligning}
Valerio Basile and Johan Bos. 2013.
\newblock \href {https://www.aclweb.org/anthology/W13-2101} {Aligning formal
  meaning representations with surface strings for wide-coverage text
  generation}.
\newblock In \emph{Proceedings of the 14th {E}uropean Workshop on Natural
  Language Generation}, pages 1--9, Sofia, Bulgaria. Association for
  Computational Linguistics.

\bibitem[{Bonial et~al.(2011)Bonial, Corvey, Palmer, Petukhova, and
  Bunt}]{Bonial:11}
Claire Bonial, William~J. Corvey, Martha Palmer, Volha Petukhova, and Harry
  Bunt. 2011.
\newblock A hierarchical unification of {LIRICS} and {VerbNet} semantic roles.
\newblock In \emph{Proceedings of the 5th {IEEE} International Conference on
  Semantic Computing ({ICSC} 2011)}, pages 483--489.

\bibitem[{Bos(2008)}]{step2008:boxer}
Johan Bos. 2008.
\newblock \href {http://www.aclweb.org/anthology/W08-2222} {{Wide-Coverage
  Semantic Analysis with Boxer}}.
\newblock In Johan Bos and Rodolfo Delmonte, editors, \emph{Semantics in Text
  Processing. STEP 2008 Conference Proceedings}, volume~1 of \emph{Research in
  Computational Semantics}, pages 277--286. College Publications.

\bibitem[{Bos et~al.(2017)Bos, Basile, Evang, Venhuizen, and Bjerva}]{GMB:2017}
Johan Bos, Valerio Basile, Kilian Evang, Noortje Venhuizen, and Johannes
  Bjerva. 2017.
\newblock \href {http://www.springer.com/la/book/9789402408799} {{The Groningen
  Meaning Bank}}.
\newblock In Nancy Ide and James Pustejovsky, editors, \emph{Handbook of
  Linguistic Annotation}. Springer Netherlands.

\bibitem[{van~der Goot et~al.(2020)van~der Goot, {\"U}st{\"u}n, Ramponi, and
  Plank}]{vandergoot-etal-2020-machamp}
Rob van~der Goot, Ahmet {\"U}st{\"u}n, Alan Ramponi, and Barbara Plank. 2020.
\newblock \href {http://arxiv.org/abs/2005.14672} {Massive choice, ample tasks
  (machamp): A toolkit for multi-task learning in nlp}.

\bibitem[{Haji{\v{c}} et~al.(2012)Haji{\v{c}}, Haji{\v{c}}ov{\'a},
  Panevov{\'a}, Sgall, Bojar, Cinkov{\'a}, Fu{\v{c}}{\'\i}kov{\'a},
  Mikulov{\'a}, Pajas, Popelka, Semeck{\'y}, {\v{S}}indlerov{\'a},
  {\v{S}}t{\v{e}}p{\'a}nek, Toman, Ure{\v{s}}ov{\'a}, and
  {\v{Z}}abokrtsk{\'y}}]{hajic-etal-2012-announcing}
Jan Haji{\v{c}}, Eva Haji{\v{c}}ov{\'a}, Jarmila Panevov{\'a}, Petr Sgall,
  Ond{\v{r}}ej Bojar, Silvie Cinkov{\'a}, Eva Fu{\v{c}}{\'\i}kov{\'a}, Marie
  Mikulov{\'a}, Petr Pajas, Jan Popelka, Ji{\v{r}}{\'\i} Semeck{\'y}, Jana
  {\v{S}}indlerov{\'a}, Jan {\v{S}}t{\v{e}}p{\'a}nek, Josef Toman, Zde{\v{n}}ka
  Ure{\v{s}}ov{\'a}, and Zden{\v{e}}k {\v{Z}}abokrtsk{\'y}. 2012.
\newblock \href
  {http://www.lrec-conf.org/proceedings/lrec2012/pdf/510_Paper.pdf} {Announcing
  {P}rague {C}zech-{E}nglish {D}ependency {T}reebank 2.0}.
\newblock In \emph{Proceedings of the Eighth International Conference on
  Language Resources and Evaluation ({LREC}'12)}, pages 3153--3160, Istanbul,
  Turkey. European Language Resources Association (ELRA).

\bibitem[{Heim(1982)}]{Heim1982}
Irene Heim. 1982.
\newblock \emph{The Semantics of Definite and Indefinite Noun Phrases}.
\newblock Ph.D. thesis, University of Massachusetts, Amherst.

\bibitem[{Hershcovich et~al.(2019)Hershcovich, Aizenbud, Choshen, Sulem,
  Rappoport, and Abend}]{hershcovich-etal-2019-semeval}
Daniel Hershcovich, Zohar Aizenbud, Leshem Choshen, Elior Sulem, Ari Rappoport,
  and Omri Abend. 2019.
\newblock \href {https://doi.org/10.18653/v1/S19-2001} {{S}em{E}val-2019 task
  1: Cross-lingual semantic parsing with {UCCA}}.
\newblock In \emph{Proceedings of the 13th International Workshop on Semantic
  Evaluation}, pages 1--10, Minneapolis, Minnesota, USA. Association for
  Computational Linguistics.

\bibitem[{Kamp(1981)}]{kamp81}
Hans Kamp. 1981.
\newblock A theory of truth and semantic representation.
\newblock In J.~A.~G. Groenendijk, T.~M.~V. Janssen, and M.~B.~J. Stokhof,
  editors, \emph{Formal Methods in the Study of Language}, volume~1, pages
  277--322. Mathematisch Centrum, Amsterdam.

\bibitem[{Kamp and Reyle(1993)}]{kampreyle:drt}
Hans Kamp and Uwe Reyle. 1993.
\newblock \emph{From Discourse to Logic; An Introduction to Modeltheoretic
  Semantics of Natural Language, Formal Logic and DRT}.
\newblock Kluwer, Dordrecht.

\bibitem[{Liu et~al.(2018)Liu, Cohen, and Lapata}]{liu-etal-2018-discourse}
Jiangming Liu, Shay~B. Cohen, and Mirella Lapata. 2018.
\newblock \href {https://doi.org/10.18653/v1/P18-1040} {Discourse
  representation structure parsing}.
\newblock In \emph{Proceedings of the 56th Annual Meeting of the Association
  for Computational Linguistics (Volume 1: Long Papers)}, pages 429--439,
  Melbourne, Australia. Association for Computational Linguistics.

\bibitem[{May(2016)}]{may-2016-semeval}
Jonathan May. 2016.
\newblock \href {https://doi.org/10.18653/v1/S16-1166} {{S}em{E}val-2016 task
  8: Meaning representation parsing}.
\newblock In \emph{Proceedings of the 10th International Workshop on Semantic
  Evaluation ({S}em{E}val-2016)}, pages 1063--1073, San Diego, California.
  Association for Computational Linguistics.

\bibitem[{May and Priyadarshi(2017)}]{may-priyadarshi-2017-semeval}
Jonathan May and Jay Priyadarshi. 2017.
\newblock \href {https://doi.org/10.18653/v1/S17-2090} {{S}em{E}val-2017 task
  9: {A}bstract {M}eaning {R}epresentation parsing and generation}.
\newblock In \emph{Proceedings of the 11th International Workshop on Semantic
  Evaluation ({S}em{E}val-2017)}, pages 536--545, Vancouver, Canada.
  Association for Computational Linguistics.

\bibitem[{Miller(1995)}]{miller1995wordnet}
George~A Miller. 1995.
\newblock Wordnet: a lexical database for english.
\newblock \emph{Communications of the ACM}, 38(11):39--41.

\bibitem[{van Noord et~al.(2018{\natexlab{a}})van Noord, Abzianidze, Haagsma,
  and Bos}]{van-noord-etal-2018-evaluating}
Rik van Noord, Lasha Abzianidze, Hessel Haagsma, and Johan Bos.
  2018{\natexlab{a}}.
\newblock \href {https://www.aclweb.org/anthology/L18-1267} {Evaluating scoped
  meaning representations}.
\newblock In \emph{Proceedings of the Eleventh International Conference on
  Language Resources and Evaluation ({LREC}-2018)}, Miyazaki, Japan. European
  Languages Resources Association (ELRA).

\bibitem[{van Noord et~al.(2018{\natexlab{b}})van Noord, Abzianidze, Toral, and
  Bos}]{van-noord-etal-2018-exploring}
Rik van Noord, Lasha Abzianidze, Antonio Toral, and Johan Bos.
  2018{\natexlab{b}}.
\newblock \href {https://doi.org/10.1162/tacl_a_00241} {Exploring neural
  methods for parsing discourse representation structures}.
\newblock \emph{Transactions of the Association for Computational Linguistics},
  6:619--633.

\bibitem[{Oepen et~al.(2020)Oepen, Abend, Abzianidze, Bos, Haji\v{c},
  Hershcovich, Li, O'Gorman, Xue, and Zeman}]{Oep:Abe:Abz:20}
Stephan Oepen, Omri Abend, Lasha Abzianidze, Johan Bos, Jan Haji\v{c}, Daniel
  Hershcovich, Bin Li, Tim O'Gorman, Nianwen Xue, and Daniel Zeman. 2020.
\newblock {MRP}~2020: {T}he {S}econd {S}hared {T}ask on {C}ross-framework and
  {C}ross-{L}ingual {M}eaning {R}epresentation {P}arsing.
\newblock In \emph{Proceedings of the {CoNLL} 2020 {S}hared {T}ask:
  {C}ross-{F}ramework {M}eaning {R}epresentation {P}arsing}, pages
  \pages{--}{1}{22}, Online.

\bibitem[{Oepen et~al.(2019)Oepen, Abend, Haji\v{c}, Hershcovich, Kuhlmann,
  O'Gorman, Xue, Chun, Straka, and Ure\v{s}ov{\'a}}]{Oep:Abe:Haj:19}
Stephan Oepen, Omri Abend, Jan Haji\v{c}, Daniel Hershcovich, Marco Kuhlmann,
  Tim O'Gorman, Nianwen Xue, Jayeol Chun, Milan Straka, and Zde\v{n}ka
  Ure\v{s}ov{\'a}. 2019.
\newblock {MRP}~2019: {C}ross-framework {M}eaning {R}epresentation {P}arsing.
\newblock In \emph{Proceedings of the Shared Task on Cross-Framework Meaning
  Representation Parsing at the 2019 {C}onference on {C}omputational {N}atural
  {L}anguage {L}earning}, pages \pages{--}{1}{27}, Hong Kong, China.

\bibitem[{Oepen et~al.(2015)Oepen, Kuhlmann, Miyao, Zeman, Cinkov{\'a},
  Flickinger, Haji{\v{c}}, and Ure{\v{s}}ov{\'a}}]{oepen-etal-2015-semeval}
Stephan Oepen, Marco Kuhlmann, Yusuke Miyao, Daniel Zeman, Silvie Cinkov{\'a},
  Dan Flickinger, Jan Haji{\v{c}}, and Zde{\v{n}}ka Ure{\v{s}}ov{\'a}. 2015.
\newblock \href {https://doi.org/10.18653/v1/S15-2153} {{S}em{E}val 2015 task
  18: Broad-coverage semantic dependency parsing}.
\newblock In \emph{Proceedings of the 9th International Workshop on Semantic
  Evaluation ({S}em{E}val 2015)}, pages 915--926, Denver, Colorado. Association
  for Computational Linguistics.

\bibitem[{Oepen et~al.(2014)Oepen, Kuhlmann, Miyao, Zeman, Flickinger,
  Haji{\v{c}}, Ivanova, and Zhang}]{oepen-etal-2014-semeval}
Stephan Oepen, Marco Kuhlmann, Yusuke Miyao, Daniel Zeman, Dan Flickinger, Jan
  Haji{\v{c}}, Angelina Ivanova, and Yi~Zhang. 2014.
\newblock \href {https://doi.org/10.3115/v1/S14-2008} {{S}em{E}val 2014 task 8:
  Broad-coverage semantic dependency parsing}.
\newblock In \emph{Proceedings of the 8th International Workshop on Semantic
  Evaluation ({S}em{E}val 2014)}, pages 63--72, Dublin, Ireland. Association
  for Computational Linguistics.

\bibitem[{Oepen and L{\o}nning(2006)}]{oepen-lonning-2006-discriminant}
Stephan Oepen and Jan~Tore L{\o}nning. 2006.
\newblock \href {http://www.lrec-conf.org/proceedings/lrec2006/pdf/364_pdf.pdf}
  {Discriminant-based {MRS} banking}.
\newblock In \emph{Proceedings of the Fifth International Conference on
  Language Resources and Evaluation ({LREC}{'}06)}, Genoa, Italy. European
  Language Resources Association (ELRA).

\bibitem[{Parsons(1990)}]{Parsons1990-PAREIT}
Terence Parsons. 1990.
\newblock \emph{Events in the Semantics of English: A Study in Subatomic
  Semantics}.
\newblock MIT Press.

\bibitem[{Power(1999)}]{Power99-enlgw}
Richard Power. 1999.
\newblock \href {http://www.itri.bton.ac.uk/~Richard.Power/ewnlg-drt.ps}
  {Controlling logical scope in text generation}.
\newblock In \emph{Proceedings of the 7th. European Workshop on Natural
  Language Generation ({EWNLG'99})}, pages 1--9, Toulouse.

\bibitem[{Van~der Sandt(1992)}]{van_der_sandt:92}
Rob~A. Van~der Sandt. 1992.
\newblock \href {https://doi.org/10.1093/jos/9.4.333} {Presupposition
  projection as anaphora resolution}.
\newblock \emph{Journal of Semantics}, 9(4):333--377.

\bibitem[{Sgall et~al.(1986)Sgall, Haji\v{c}ov\'{a}, and
  Panevov\'{a}}]{sgallhp:1986}
Petr Sgall, Eva Haji\v{c}ov\'{a}, and Jarmila Panevov\'{a}. 1986.
\newblock \emph{{The Meaning of the Sentence and Its Semantic and Pragmatic
  Aspects}}.
\newblock Academia/Reidel Publishing Company, Prague, Czech Republic/Dordrecht,
  Netherlands.

\bibitem[{Venhuizen et~al.(2013)Venhuizen, Bos, and
  Brouwer}]{venhuizen-etal-2013-parsimonious}
Noortje~J. Venhuizen, Johan Bos, and Harm Brouwer. 2013.
\newblock \href {https://www.aclweb.org/anthology/W13-0122} {Parsimonious
  semantic representations with projection pointers}.
\newblock In \emph{Proceedings of the 10th International Conference on
  Computational Semantics ({IWCS} 2013) {--} Long Papers}, pages 252--263,
  Potsdam, Germany. Association for Computational Linguistics.

\bibitem[{Zeman and Haji\v{c}(2020)}]{Zem:Haj:20}
Daniel Zeman and Jan Haji\v{c}. 2020.
\newblock {FGD} at {MRP}~2020: {P}rague {T}ectogrammatical {G}raphs.
\newblock In \emph{Proceedings of the {CoNLL} 2020 {S}hared {T}ask:
  {C}ross-{F}ramework {M}eaning {R}epresentation {P}arsing}, pages
  \pages{--}{33}{39}, Online.

\bibitem[{Žabokrtský et~al.(2020)Žabokrtský, Zeman, and
  Ševčíková}]{zabokrtsky-et-al-2020}
Zdeněk Žabokrtský, Daniel Zeman, and Magda Ševčíková. 2020.
\newblock \href {https://doi.org/10.1162/coli\_a\_00385} {Sentence meaning
  representations across languages: What can we learn from existing
  frameworks?}
\newblock \emph{Computational Linguistics}, 0(0):605--665.

\end{thebibliography}

\end{document}